\documentclass[runningheads]{llncs}
\usepackage{graphicx}
\usepackage[T1]{fontenc}
\usepackage[latin9]{inputenc}
\usepackage{graphicx}
\usepackage{babel}
\usepackage[]{algorithm2e}
\usepackage{tikz,float}
\usepackage{pgfplots}
\def \hfillx {\hspace*{-\textwidth} \hfill}
\usepackage{amsmath}

\usetikzlibrary{positioning, shapes.geometric,arrows.meta}

\usepackage[unicode=true, bookmarks=true,bookmarksnumbered=true,bookmarksopen=true,bookmarksopenlevel=5, breaklinks=false,pdfborder={0 0 0},pdfborderstyle={},backref=page,colorlinks=true] {hyperref}

\begin{document}
\title{An Analytical Study of Covid-19 Dataset using Graph-Based Clustering Algorithms\thanks{Supported by NIT Trichy Computer Applications Department. \Large\bfseries}}
%
%\titlerunning{Abbreviated paper title}
% If the paper title is too long for the running head, you can set
% an abbreviated paper title here
%

\author{Mamata Das\inst{1}\orcidID{0000-0002-5106-5571} \and
	PJA Alphonse \inst{2}\orcidID{0000-0003-2899-5911} \and
	Selvakumar K \inst{3}\orcidID{0000-0002-8376-6030}
}

%\author{Mamata Das\inst{1}\orcidID{0000-0002-5106-5571} \and
%PJA Alphonse\inst{2}\orcidID{0000-0003-2899-5911} \and
%Selvakumar K \inst{3}\orcidID{0000-0002-8376-6030}
%
\authorrunning{M. Das et al.}
% First names are abbreviated in the running head.
% If there are more than two authors, 'et al.' is used.
%
\institute{NIT Trichy, Tamil Nadu, 620015, India \\
%\email{\{dasmamata.india@gmail.com, {\{alphonse,kselvakumar\}@nitt.edu }
\email{dasmamata.india@gmail.com,\{alphonse, kselvakumar\}@nitt.edu} }
%\email{lncs@springer.com}
%
\maketitle              % typeset the header of the contribution
\begin{abstract}
	COrona VIrus Disease abbreviated as COVID-19 is a novel virus which is initially identified in Wuhan of China in December of 2019 and now this deadly disease has spread all over the world. According to World Health Organization (WHO), a total of 3,124,905 people died from 2019 to 2021, April. In this case, many methods, AI base techniques, and machine learning algorithms have been researched and are being used to save people from this pandemic. The SARS-CoV and the  2019-nCoV, SARS-CoV-2 virus invade our bodies, causing some differences in the structure of cell proteins. Protein-protein interaction (PPI) is an essential process in our cells and plays a very important role in the development of medicines and gives ideas about the disease. In this study, we performed clustering on PPI networks generated from 92 genes of the Covi-19 dataset. We have used three graph-based clustering algorithms to give intuition to the analysis of clusters.

\end{abstract}
\section{\large\bfseries Introduction}

Clustering is a task where similar objects are grouped into the same groups and others into different groups. In statistical data analysis, clustering is used as a common technique in different fields. Some uses are in machine learning, computer graphics, bioinformatics, information retrieval, image analysis, and pattern recognition. There are so many clustering algorithms where data points are coming from Gaussian distribution like k-means \cite{jain1988algorithms}\cite{selvakumar2015enhanced}. Togate the high quality result we need to know some prior knowledge about clusterings like parameter or threshold value, the initial number of clusters, etc. To solve this problem, we can think of an algorithm that belongs to graph theory i.e., graph-based clustering algorithms where problems are categorized as Undirected Graph. In graph-based clustering, data are transformed into a graph representation \cite{ROY2017361}. Data points are the vertices of the graph to be clustered. The edges between the data points or the node are weighted by their similarity. Usually, the model does not require any prior knowledge however users provide some parameters with value. Graph-based clustering methods are broadly used in biomedical and biological studies to determine the relationship among the objects \cite{YupingCancer}. 
\clearpage

%An efficient method is proposed by \cite{enright2002efficient} called TRIBE-MCL. It is based on Markov chain theory \cite{Anton} and applied on protein sequence like a way, protein represented as a node and edge of the graph containing a weight is scientifically computed similarity values.

We have used three interesting and efficient graph-based clustering algorithms in our experiment namely Markov Clustering Algorithm (MCL) as Algorithm one ($A_1$), Regularized Markov Clustering Algorithm (RMCL) as Algorithm two ($A_2$), and MCL with the variable inflation rate Algorithm three ($A_3$). We have used both Covid-19 real data as well as synthetic datasets to evaluate and compare different graph-based algorithms and analyses. The result of the analysis of all the clusters is shown in section 4. It shows that the performance of $A_1$ is adequate and $A_2$ performs better than $A_3$.

The rest of this paper is indexed as follows. Section 2 presents an overview of the literature survey of previous works on graph-based clustering. Section 3 presents the materials and methods. Section 4 shows the performance of experimental results. Section 5 presents discussions and conclusions.

\section{Related Work}\vspace{2mm}
Two novel graph attacks are presented in the network-level graph-based detection system \cite{chen2017practical}. They have highlighted their work in the adversarial machine learning area where there are not many graph-based clustering techniques out there yet. Three clustering methods have been used namely the Community Detection method, Spectral Methods, and node2vec method, and showed the selection criteria of hyperparameters for each of the three graph methods. The result of the study indicates that a real-world system that uses several popular graph-based modeling techniques can break by generic attacks and the cost of the attacks is minimum.

A comparative study has been proposed on Markov chain correlation and shows that their method defeats the K-means clustering method \cite{1425235}. They have taken gene expression data for experiments and used Dunn Index for Evaluation purposes.

Authors have proposed a beautiful study of the PPI network on candidate genes of the Schizophrenia dataset \cite{Rizki_2021}. They have been implemented and simulated RMCL graph-based clustering algorithm and the result is compared with the MCL algorithm on the same parameters. %The conclusion of the research was the relation between protein clusters of candidate genes.

An efficient method is proposed by \cite{enright2002efficient} called TRIBE-MCL. It is based on Markov chain theory \cite{Anton} and applied on protein sequence like a way, protein represented as a node and edge of the graph containing a weight is scientifically computed similarity values.

Authors have used Stepping-stone type RMCL on Japanese associative concept dictionary and got a satisfactory level of performance than the Markov clustering algorithms generated network \cite{jung2006recurrent}. They have summarized the problems of MCL algorithms and proposed a Stepping-stone type algorithm of RMCL algorithms as an extension of the MCL Algorithm.

The paper \cite{rao2021vec2gc} concerns Vec2GC, a clustering algorithm to represent text.  The approach is density-based which is works on a document or terms. They have used graphs with weight on edges and apply community detection on the objects.
\clearpage
Applications of Ant Lion Optimization (ALO) and Cuckoo Search (CS) have been discussed on protein-protein interaction for graph-based clustering. The author has used Regularized Markov clustering method on SARS-CoV-2 and the humans dataset. The results indicate that CS-RMCL interactions are more stable than ALO-RMCL interactions\cite{Rizki_2021}.

The paper \cite{behera2020graph} proposed a way to prevent covid'19 by predicting in which area Covid-19 can spread next based on geographical distance using graph-based clustering. Here, distance threshold is used to represent the connected graph, administrative as nodes, and geographic distance as a weight of the edges. Some analysis of graph-based clustering method and performance be found in \cite{foggia2007graph,Wilshchut}.% \cite{Wilshchut}

\section{Materials and Methods}\vspace{-2mm}
\subsection{\bfseries Data collection}
We have used Covid-19 datasets in our experiment and the dataset downloaded from the website Universal Protein Resource Knowledgebase (UniProtKB) which is freely available. This latest covid-19 data set can also be accessed from the link:
$ftp://ftp.uniprot.org/pub/databases/uniprot/pre\_release/. $ 
These datasets have been collected from a database in uncompressed excel format and have 92 genes(Human (64), SARS-CoV (15) and 2019-nCoV, SARS-CoV-2 (13)). We perform clustering analysis on PPI networks using three graph-based clustering algorithms. We have used STRING to construct a PPI network, a well-known functional protein association network. The gene names of data sets are shown below.\vspace{-5mm}
\subsubsection{Homo sapiens (Human)} NRP1 NRP VEGF165R, TMPRSS2 PRSS10,TLR3, SGTA SGT SGT1, TOMM70 KIAA0719 TOM70 TOMM70A, SNAP29, HLA-B HLAB, APOE, CD74 DHLAG, HLA-A HLAA, S100A8 CAGA CFAG MRP8, IL6 IFNB2, CTSL CTSL1, IL6R, HMGB1 HMG1, FURIN FUR PACE PCSK3, HLA-E HLA-6.2 HLAE, IFNAR1 IFNAR, SMPD1 ASM, ITGAL CD11A, KLRC1 NKG2A, CIITA MHC2TA, PHB PHB1, BSG UNQ6505/PRO21383, IL6ST, IFNAR2 IFNABR, IFNARB, VPS41, RAB7A RAB7, KPNA2 RCH1 SRP1, STX17, PPIA CYPA, EEF1A1 EEF1A EF1A LENG7, SMAD3 MADH3, BST2, KLRD1 CD94, IRF5,IRF3, IL17A CTLA8 IL17, LY6E 9804 RIGE SCA2 TSA1, HIF1A BHLHE78 MOP1 PASD8, TMEM41B KIAA0033, TICAM1 PRVTIRB TRIF, MPP5 PALS1, IL17RC UNQ6118/PRO20040/PRO38901, TPCN2 TPC2, DDX1, IRF7, DHX58 D11LGP2E LGP2, IL17RA IL17R, VPS39 KIAA0770 TLP VAM6, IL17F, PHB2 BAP REA, VAMP8, ACE2 UNQ868/PRO1885, IFIH1 MDA5 RH116, NLRP1 CARD7 DEFCAP KIAA0926 NAC NALP1, TMPRSS4 TMPRSS3 UNQ776/PRO1570, ARL8B ARL10C GIE1, TBK1 NAK, PIKFYVE KIAA0981 PIP5K3 \vspace{-5mm}
\subsubsection{Severe acute respiratory syndrome coronavirus (SARS-CoV)} 1a, rep 1a-1b, S 2, N 9a, M 5, 3a, 7a, 9b, E sM 4, 3b, 6, 7b, ORF14, 8b, 8a
\subsubsection{Severe acute respiratory syndrome coronavirus 2 (2019-nCoV, SARS-CoV-2)} S 2, 3a, N, rep 1a-1b, E 4, M, 7a, 8, 6, 9b, 7b, ORF10 orf10, 9c.
\clearpage
\subsection{Execution environments}
We have implemented our experimental execution on a Lenovo ThinkPad E14 Ultrabook running the Windows 10 Professional 64-bit operating system and 10th Generation Intel Core i7-10510U Processor. The clock speed of the processor is 1.8 GHz with 16G bytes DDR4 memory size. The code has been executed in Python programming language (Version 3.6) in the Jupyter Notebook of Conda environment.

\subsection{Graph-Based Clustering method }
We have used three graph-based clustering algorithms in our experiment namely Markov Clustering Algorithm, Regularized Markov Clustering Algorithm, and MCL with the variable inflation rate.

MCL algorithm proposed by Stijn Van Dongen in 2000 \cite{vandongen2000cluster} is well known as an effectual algorithm in graph-based clustering. This algorithm is very famous in bioinformatics to cluster the protein sequence data as well as gene data. In our methodology, we give an undirected graph as an input which is constructed from the PPI network, with expansion parameter $= 6$ for a random walk and inflation parameter $= 3$ for probability modification. The algorithm got the sub-cluster as output after going to the convergence stage. One downside of the MCL algorithm is extraneous clusters in the final output and which may give us an impracticable solution. To improve this situation RMCL algorithm is used.

RMCL is the modification of MCL and  developed by Parthasarathy and Satuluri in 2009 \cite{satuluri2010markov}. The main three steps are Pruning, Inflation, and Regularization. The value of the inflation parameter is given from the outside to get good results. Here, the inflation factor value is fixed. We have applied RMCL to PPI networks of covid-19 candidate genes data. 

In order to overcome the limation of the fixed inflation parameter value, MCL with the variable inflation rate is introduced. The idea behind the variable inflation value is to modify the similarity values in the columns of the similarity matrix \cite{medves2008modified} and to get clusters with high quality.

\subsection{Protein-Protein Interaction Network}
The Protein-Protein Interaction (PPI) network has been established to predict and analyze the function of protein in terms of physical interaction. The PPI network is a model that represents graphical connectivity between proteins. Though all the proteins connected with each other in the network there may also be some isolated components. We can present it as a graph where proteins are represented as nodes and interactions are represented as an edge.

We have used STRING (Search Tool for the Retrieval of Interacting Genes / Proteins) to construct a PPI network. Cytoscape platform has been used to visualize the networks. Fig. \ref{PPI_Network_G1} and Fig. \ref{PPI_Network_G2} show the PPI graph $G_1(V_1,E_1)$ and $G_2(V_2,E_2)$ obtained from the Covid-19 dataset using the STRING tool. 
\clearpage
\section{Results and Analysis}
\begin{figure}%[!htb]
	\minipage{0.50\textwidth}\begin{center}
		\centering 
	\includegraphics[clip,scale=.15]{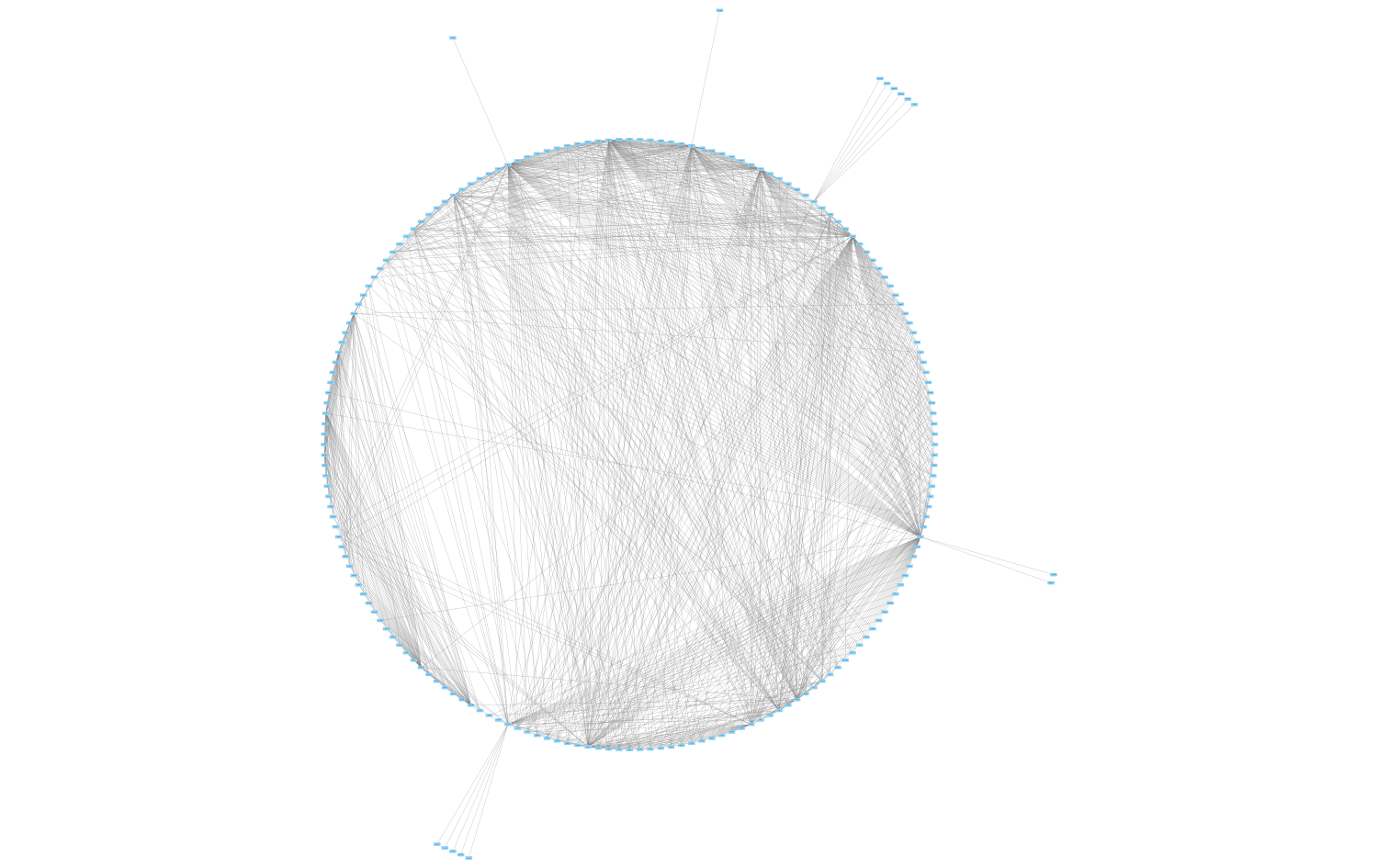}
		\end{center}\caption{PPI Network $G_1$}\label{PPI_Network_G1}
	\endminipage\hfill
	\minipage{0.50\textwidth}\begin{center}
		\centering 
	\includegraphics[clip,scale=.15]{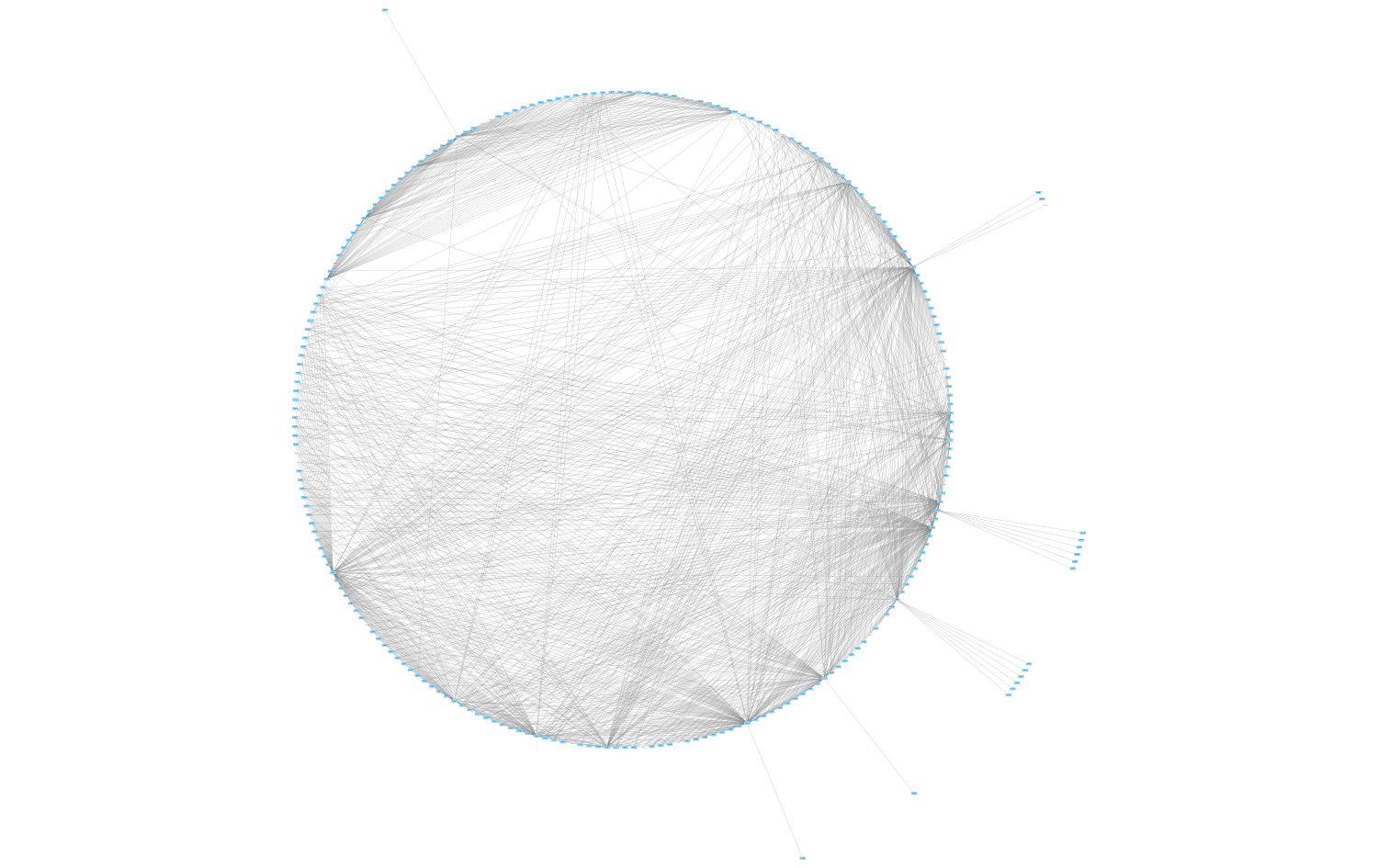}
		\end{center}\caption{PPI Network $G_2$ }\label{PPI_Network_G2}
	\endminipage
\end{figure}

\begin{figure}%[!htb]
	\minipage{0.40\textwidth}\begin{center}
		\centering \includegraphics[clip,scale=.1]{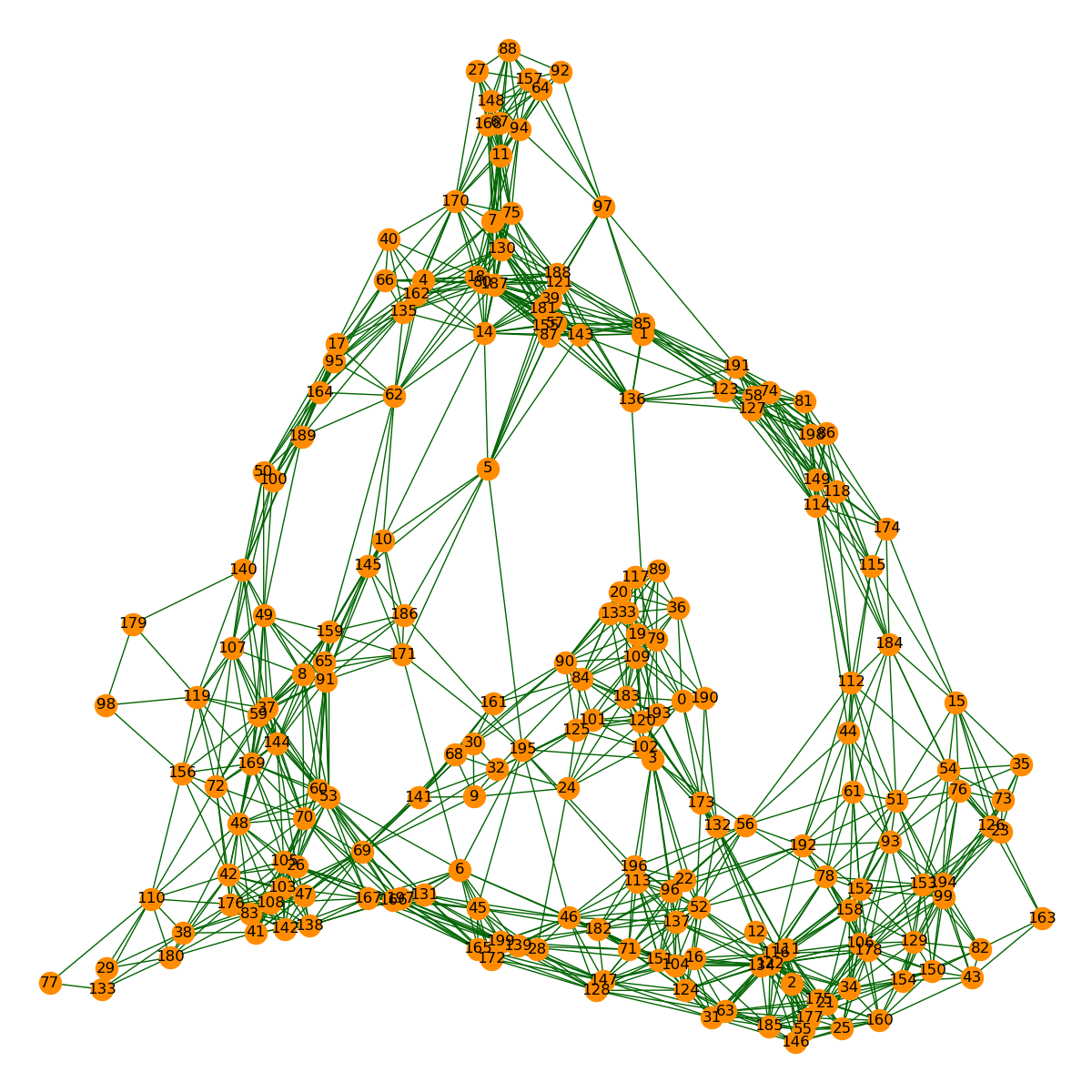}
	\end{center}
	
	\caption{Random Network $G_3$ }\label{G3}
	\endminipage\hfill
	\minipage{0.40\textwidth}
	\begin{center}
\centering 	\includegraphics[clip,scale=.1]{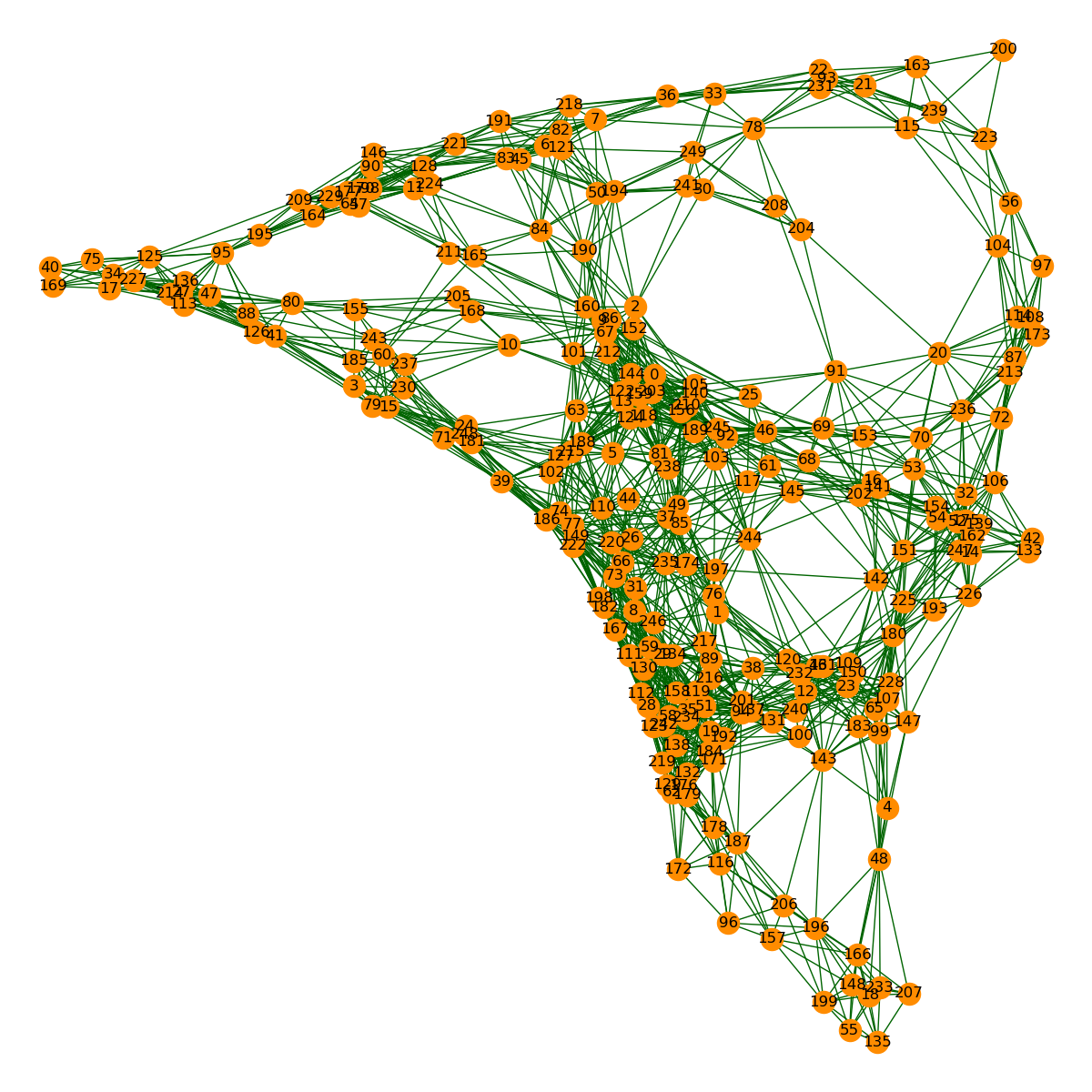}		
	\end{center}

	\caption{Random Network $G_4$  }\label{G_4}
	\endminipage
	
\end{figure}

\begin{figure}[H]%[!htb]
	\minipage{0.32\textwidth}
	\includegraphics[clip,scale=.1]{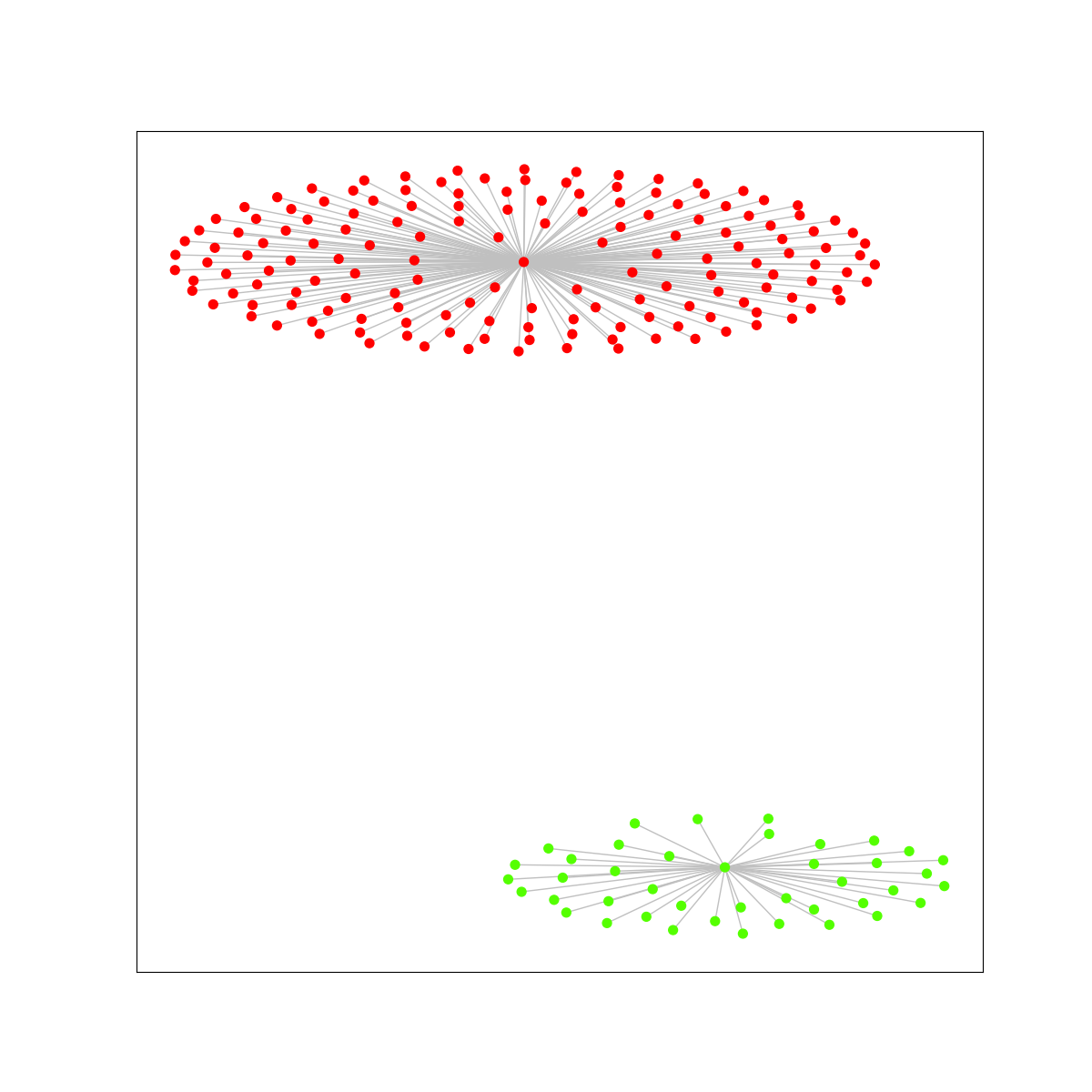}
	\caption{$G_1$ on $A_1$}\label{fig:G1A1}
	\endminipage\hfill
	\minipage{0.32\textwidth}
	\includegraphics[clip,scale=.1]{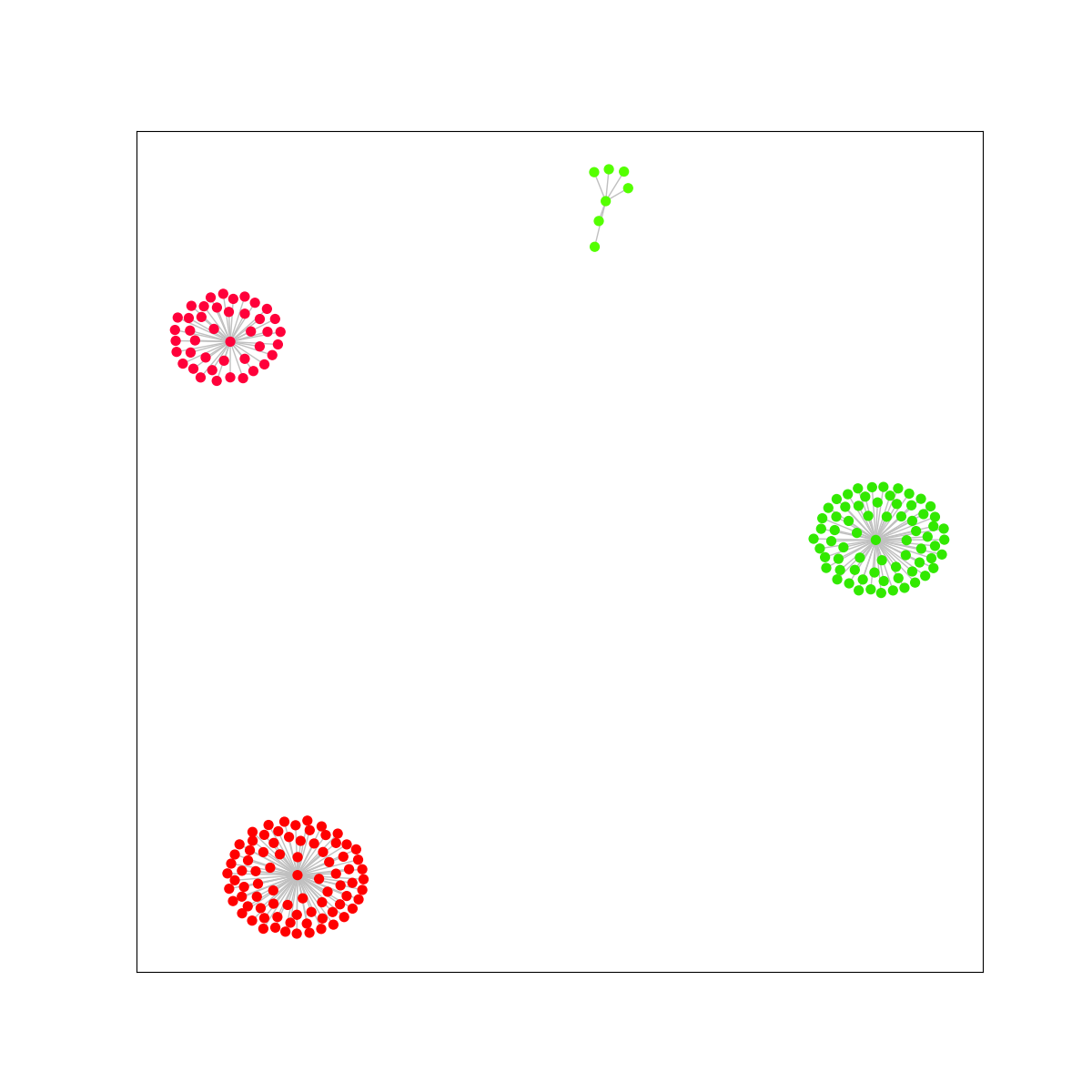}
	\caption{$G_1$ on $A_2$}\label{fig:G1A2}
	\endminipage\hfill
	\minipage{0.32\textwidth}%
	\includegraphics[clip,scale=.1]{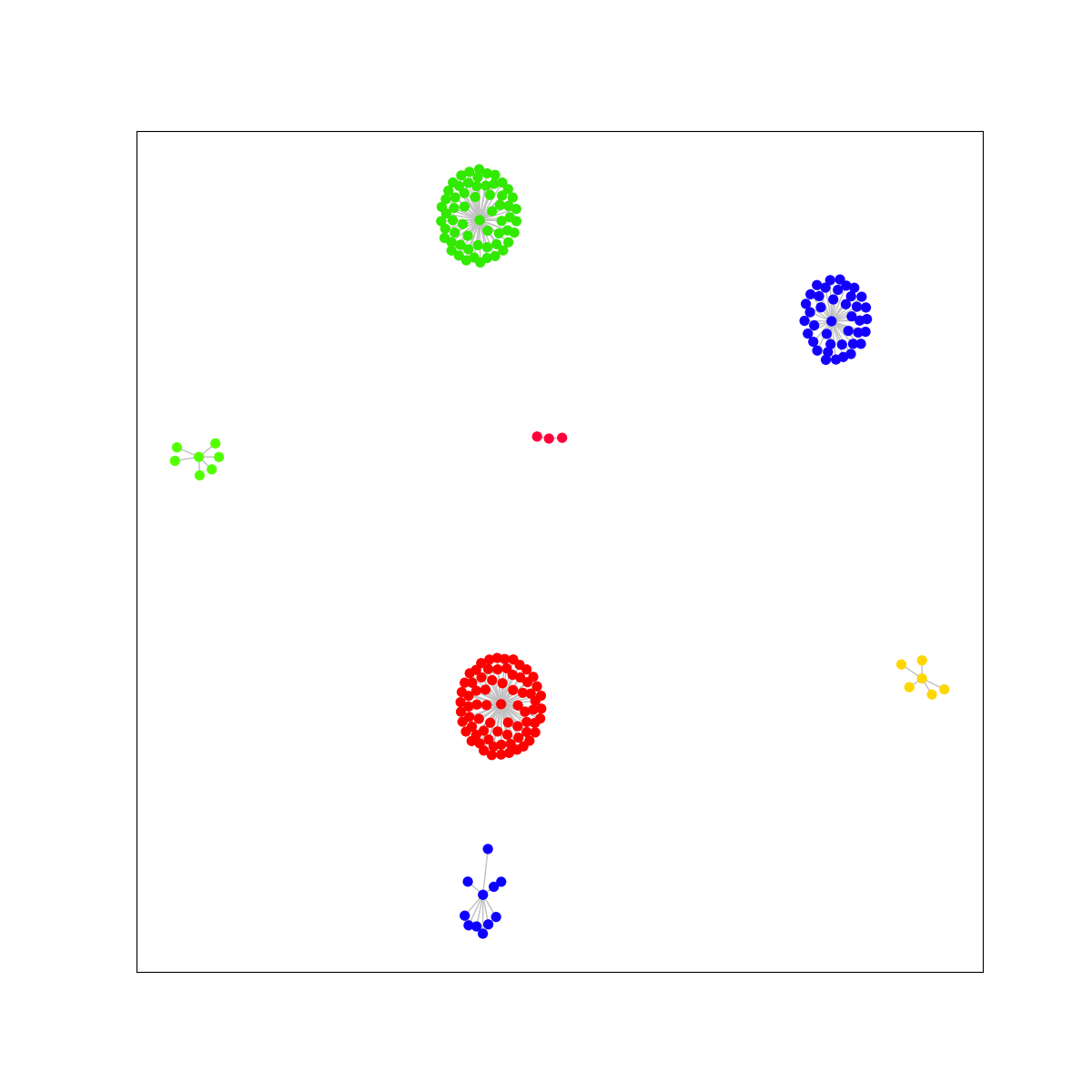}
	\caption{$G_1$ on $A_3$}\label{fig:G1A3}
	\endminipage
\end{figure}

\begin{figure}[H]%[!htb]
	\minipage{0.32\textwidth}
	\includegraphics[clip,scale=.1]{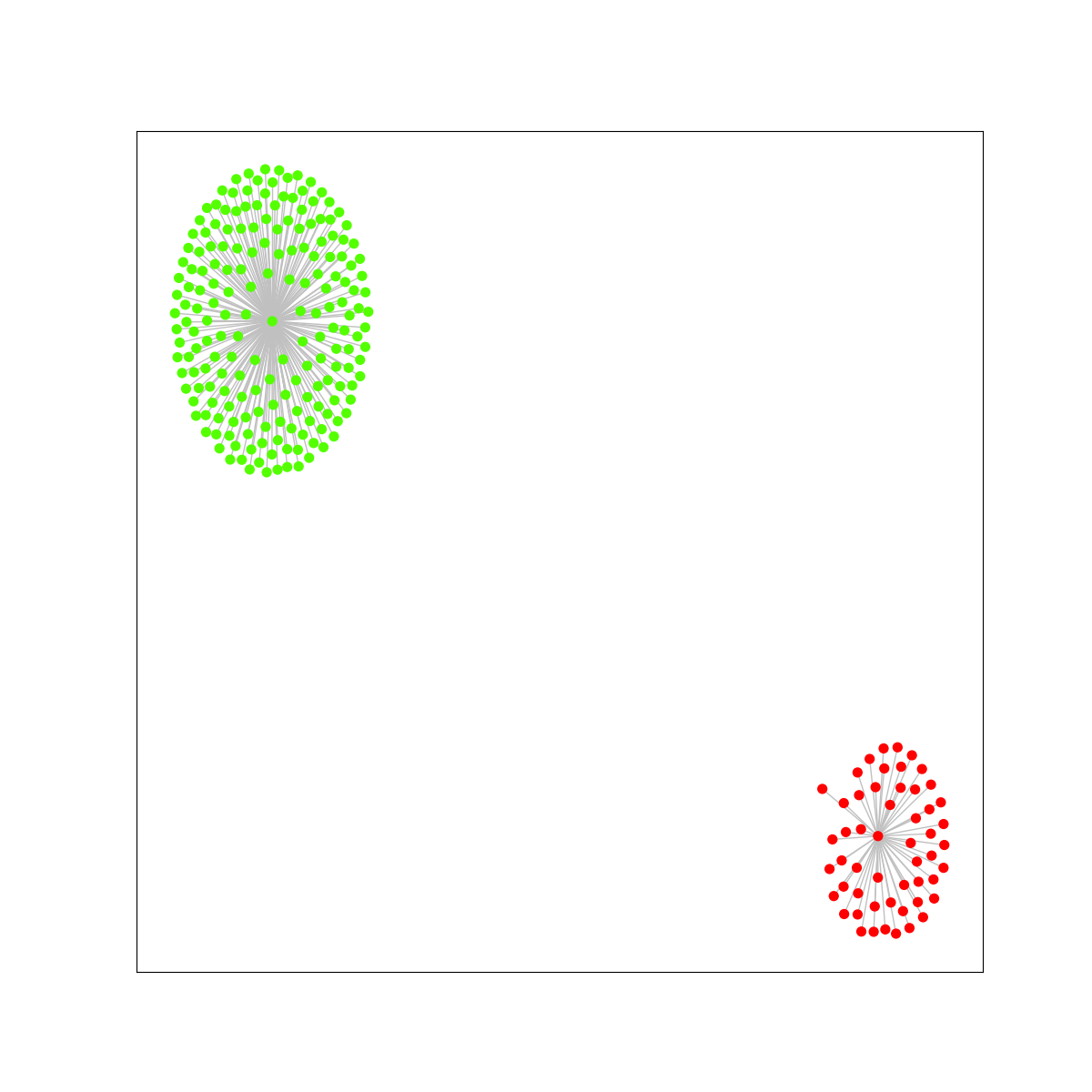}
	\caption{$G_2$ on $A_1$}\label{fig:G2A1}
	\endminipage\hfill
	\minipage{0.32\textwidth}
	\includegraphics[clip,scale=.1]{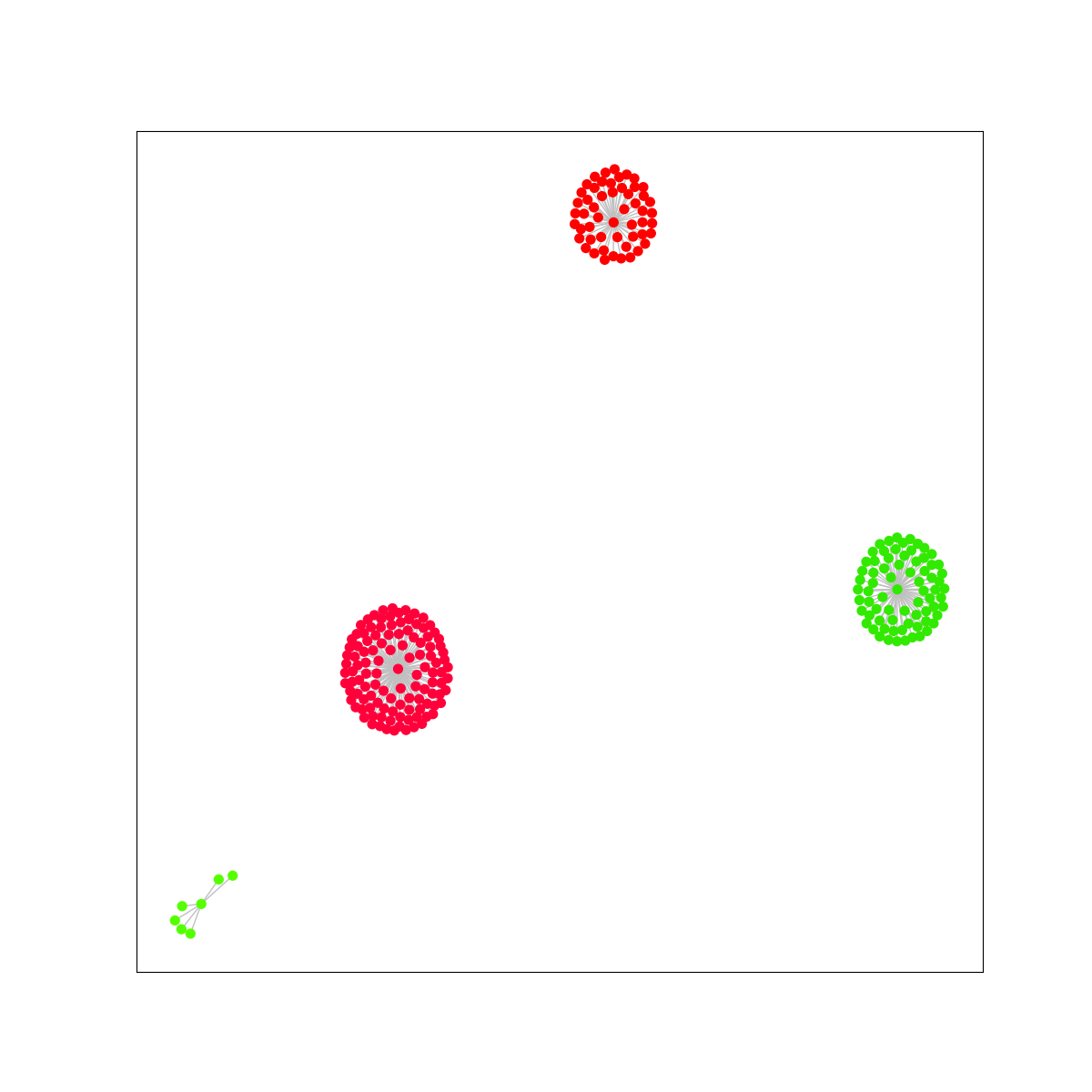}
	\caption{$G_2$ on $A_2$}\label{fig:G2A2}
	\endminipage\hfill
	\minipage{0.32\textwidth}%
	\includegraphics[clip,scale=.1]{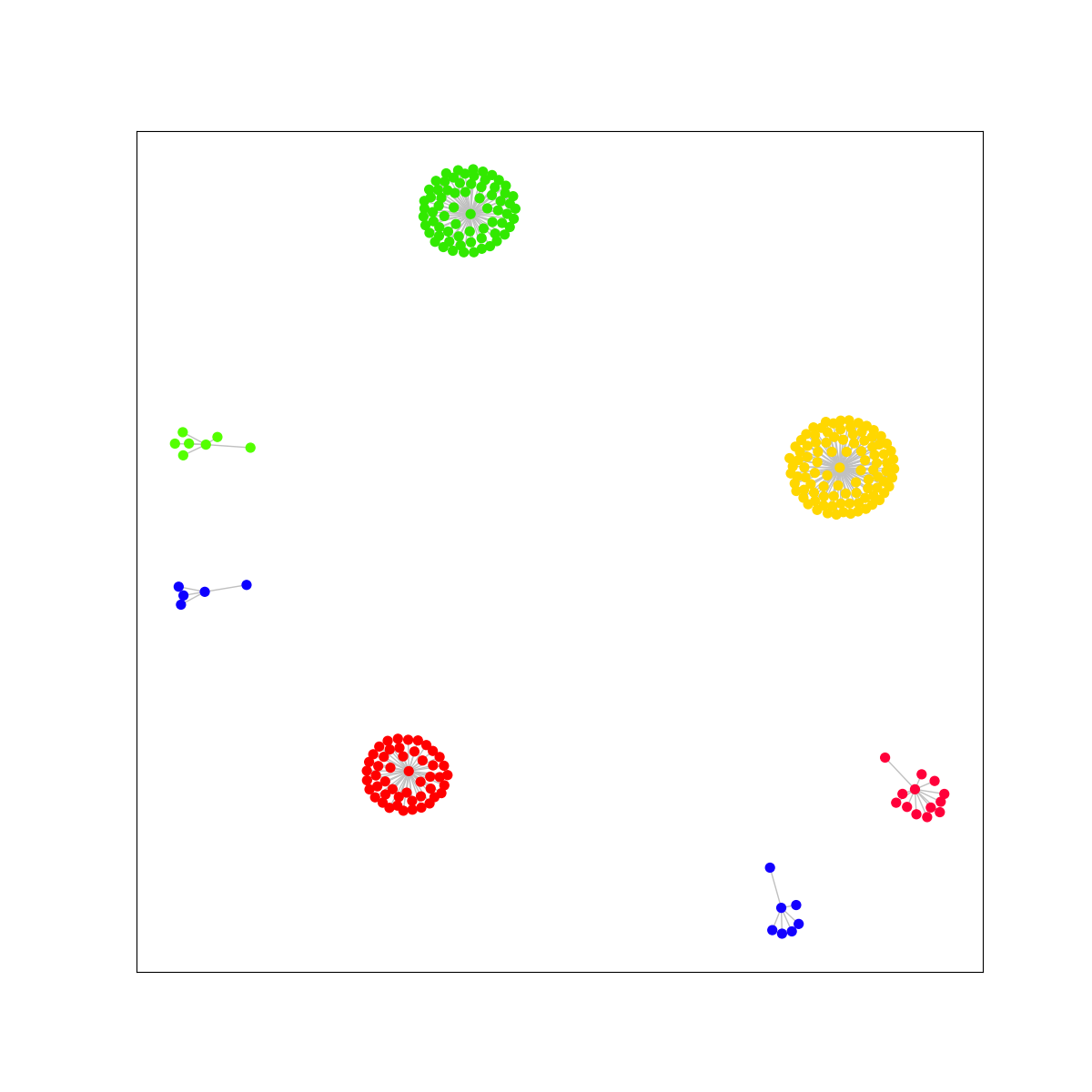}
	\caption{$G_2$ on $A_3$}\label{fig:G2A3}
	\endminipage
\end{figure}

\begin{figure}[H]%[!htb]
	\minipage{0.32\textwidth}
	\includegraphics[clip,scale=.1]{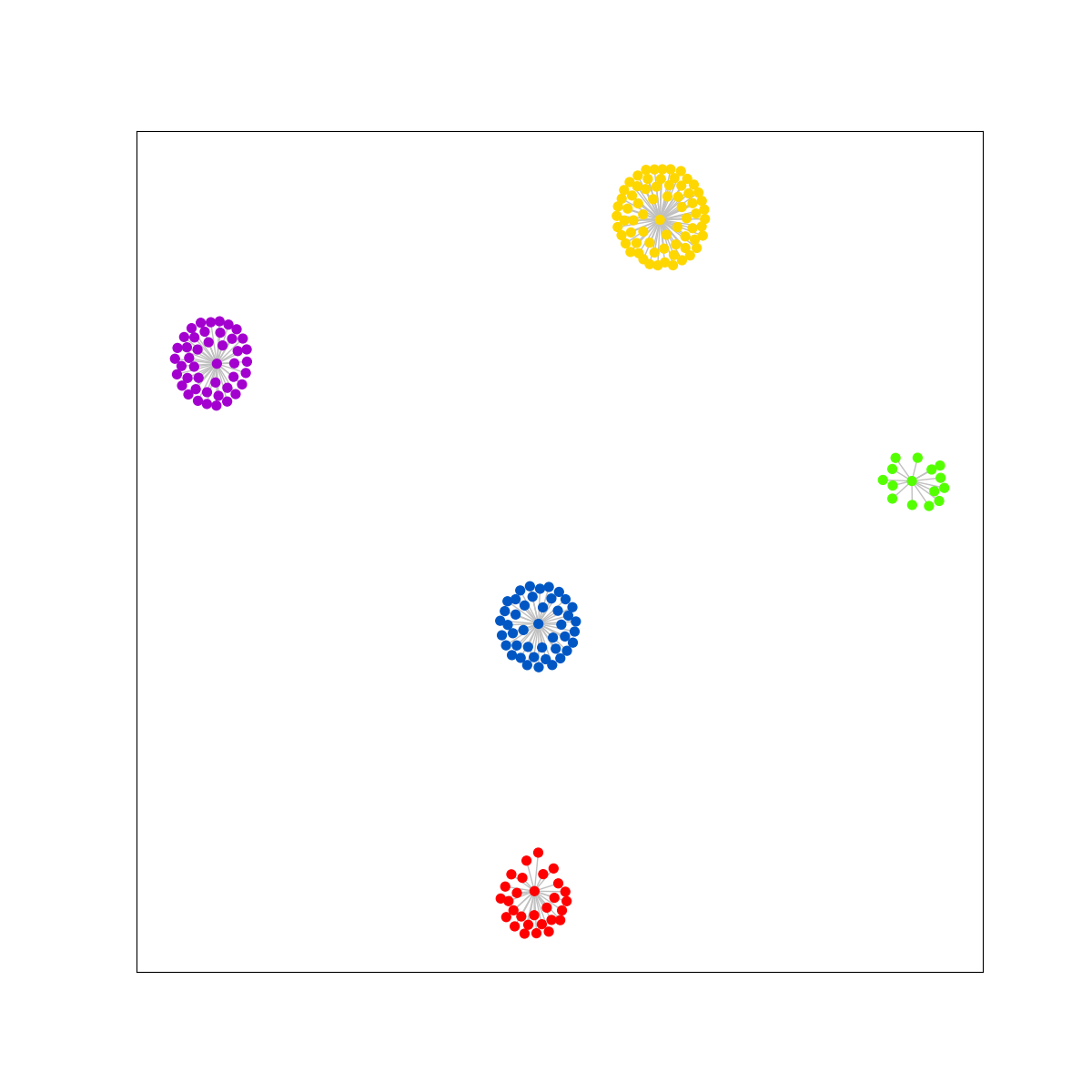}
	\caption{$G_3$ on $A_1$}\label{fig:G3A1}
	\endminipage\hfill
	\minipage{0.32\textwidth}
	\includegraphics[clip,scale=.1]{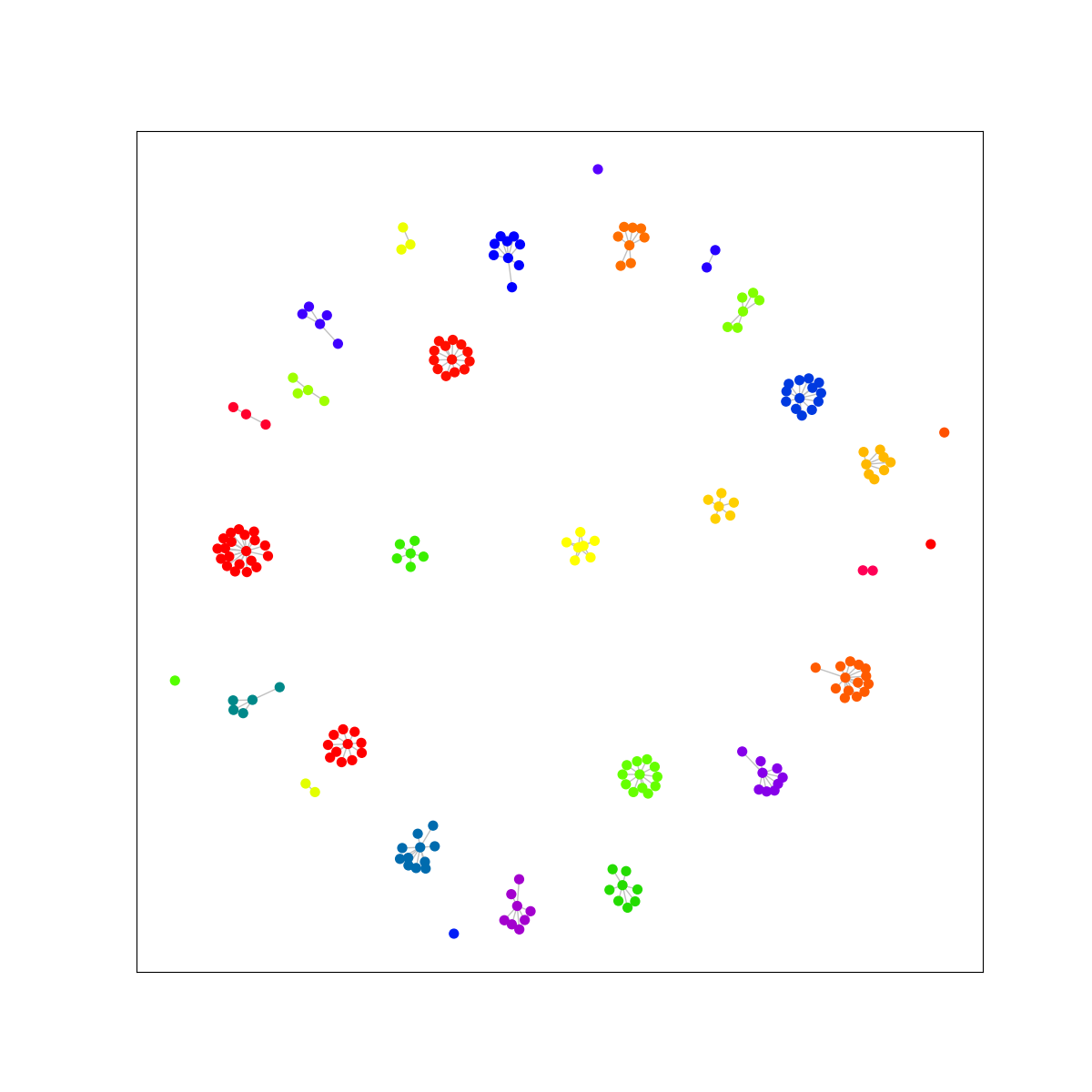}
	\caption{$G_3$ on $A_2$}\label{fig:G3A2}
	\endminipage\hfill
	\minipage{0.32\textwidth}%
	\includegraphics[clip,scale=.1]{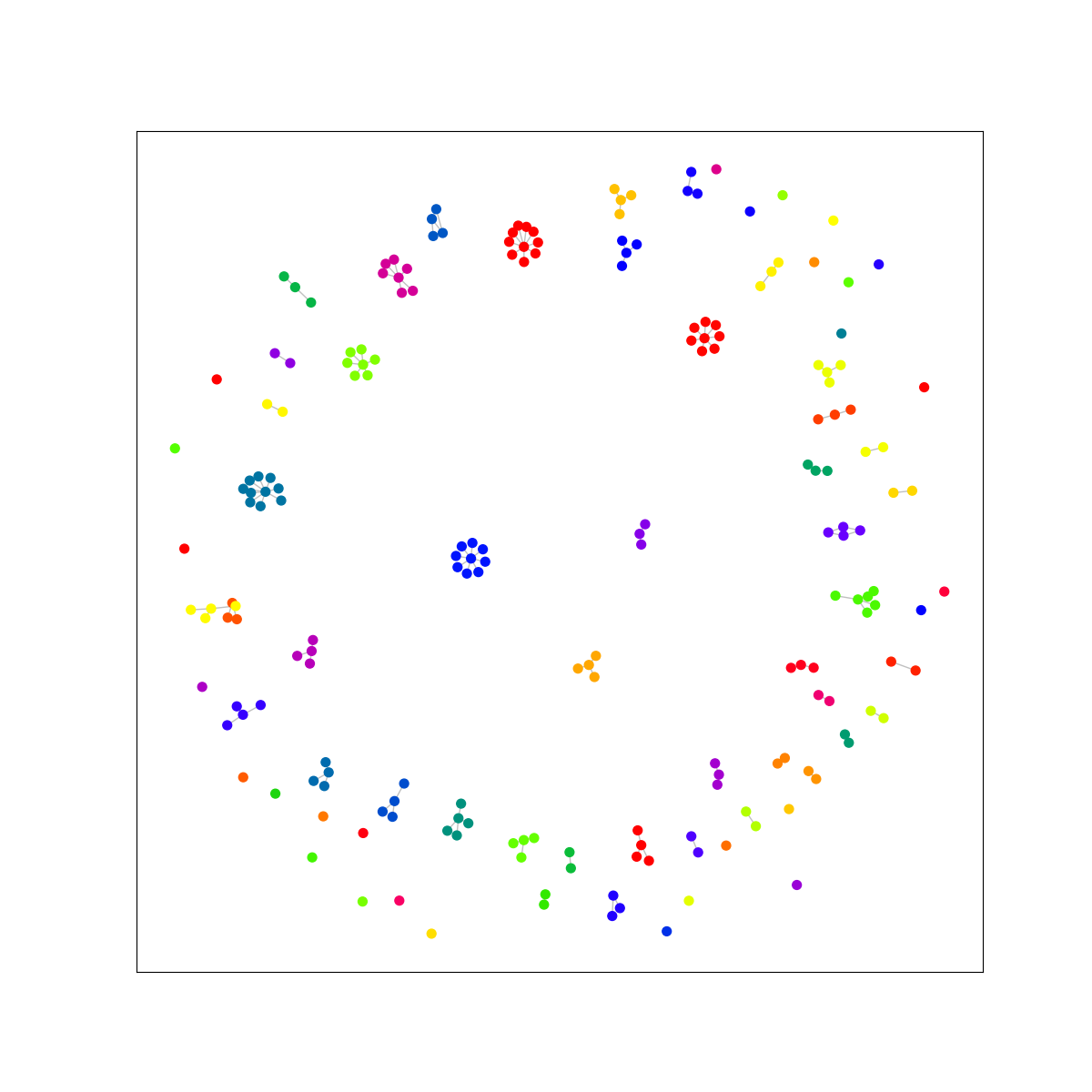}
	\caption{$G_3$ on $A_3$}\label{fig:G3A3}
	\endminipage
\end{figure}

\begin{figure}[H]%[!htb]
	\minipage{0.32\textwidth}
	\includegraphics[clip,scale=.1]{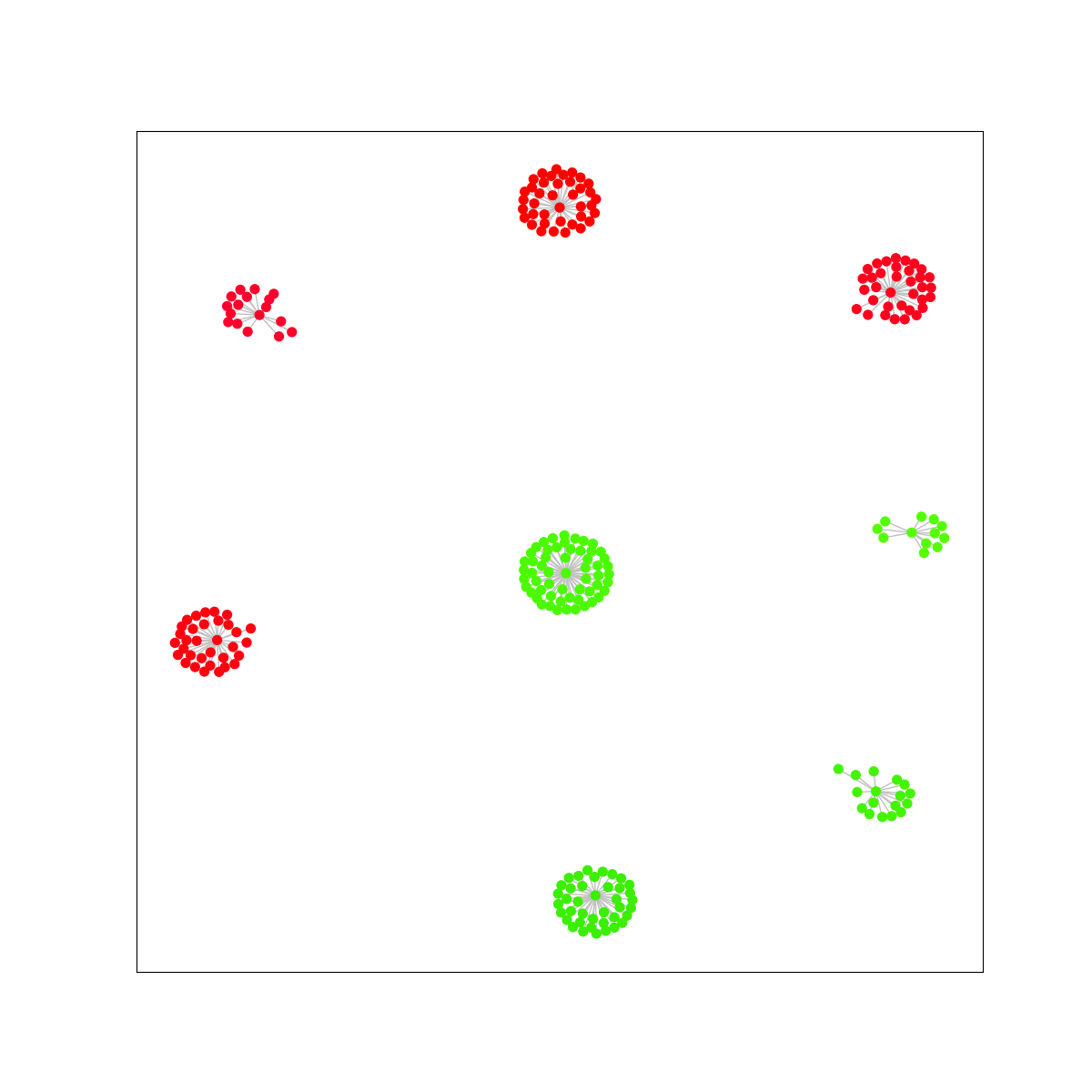}
	\caption{$G_4$ on $A_1$}\label{fig:G4A1}
	\endminipage\hfill
	\minipage{0.32\textwidth}
	\includegraphics[clip,scale=.1]{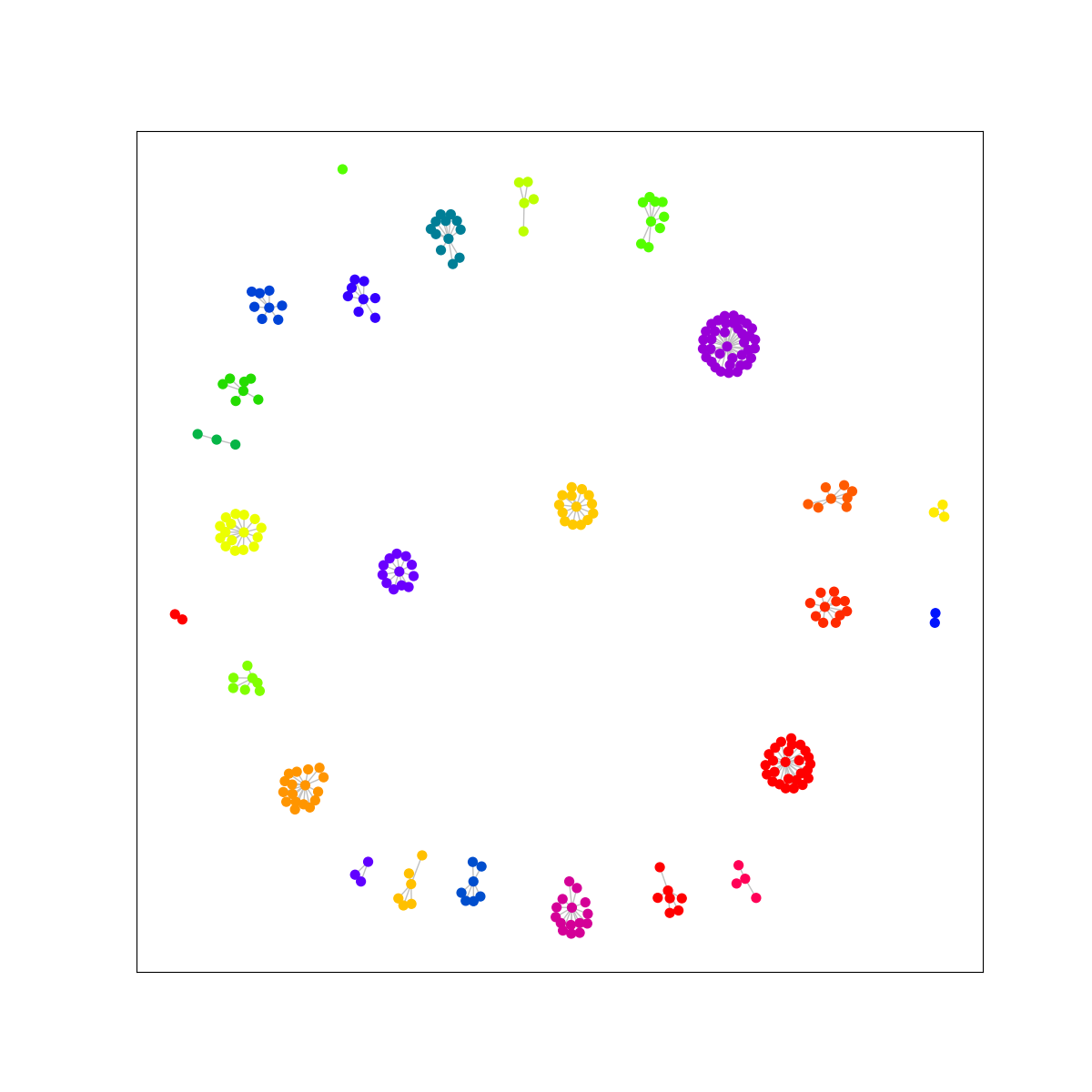}
	\caption{$G_4$ on $A_2$}\label{fig:G4A2}
	\endminipage\hfill
	\minipage{0.32\textwidth}%
	\includegraphics[clip,scale=.1]{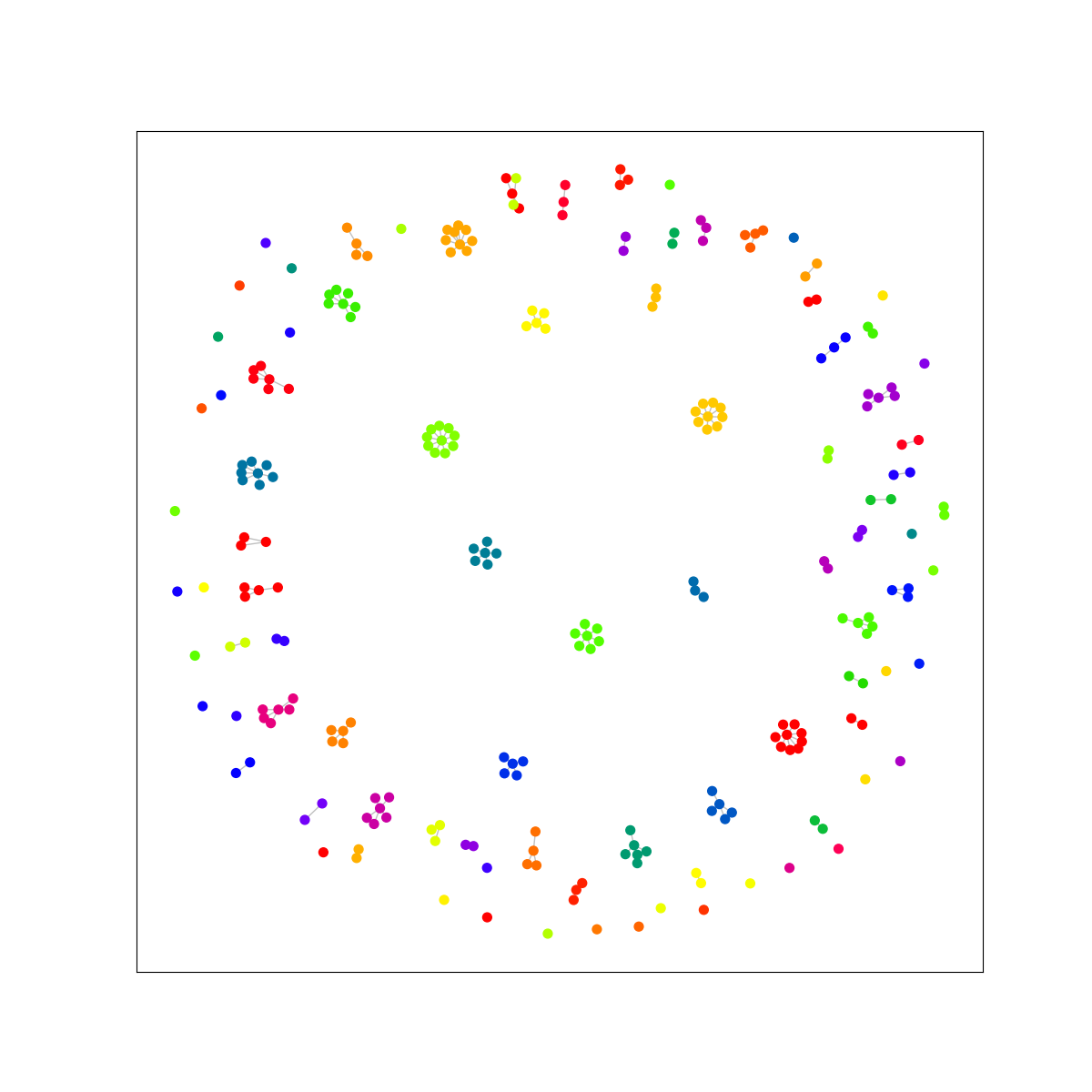}
	\caption{$G_4$ on $A_3$}\label{fig:G4A3}
	\endminipage
\end{figure}

\vspace{-4mm}
We have analyzed our results and organized them (only two PPI networks and two randomly generated graphs) by the figure from Fig. \ref{PPI_Network_G1} to Fig. \ref{DI_PPI}. We have constructed PPI networks with vertex range 150 to 250  with non-uniform increment and five randomly generated graphs with vertex range 150 to 250  with a uniform increment of 25 of the dataset. Fig. \ref{PPI_Network_G1} and Fig. \ref{PPI_Network_G2} shows the PPI graph $G_1(V_1,E_1)$, $G_2(V_2,E_2)$ and Fig. \ref{G3} and Fig. \ref{G_4} represents the random graph $G_3(V_3,E_3)$,  $G_4(V_4,E_4)$ with $V_1=204$, $V_2=254$, $V_3=200$ and $V_4=250$. Output clusters of graphs are shown in Fig. \ref{fig:G1A1} to Fig. \ref{fig:G4A3}. The iteration count versus execution time of four graphs ($G_1$, $G_2$, $G_3$, and $G_4$) are shown in Fig. \ref{E_G1} to Fig. \ref{E_G4}. We have used the sparse matrix to do the experiment. The density of the matrix has been calculated in every iteration of the experiment. Fig. \ref{S_G1} to Fig. \ref{S_G4} indicates the sparseness of the graph ($G_1$, $G_2$, $G_3$, and $G_4$). We can see the histogram of $G_3$ and $G_4$ in Fig. \ref{Histo_G3A3} and Fig. \ref{Histo_G4A3}. We have validated our clustering using the Dunn index (DI) and the quality of the clustering is very magnificent. DI obtained from the PPI network and the random graphs visualized in Fig. \ref{DI_PPI} and Fig. \ref{DI_random}.

%\section{Discussion and Conclusions}
%In this paper, we propose graph-based clustering by the help of MCL, RMCL and MCL with variable inflation rate algorithm. We have validated our clustering using DI and quality of the clustering is very magnificent. To evaluate our clustering algorithm, DI metric is interpreted as an intra-cluster and inter-cluster distance. DI in the results section shows that performance of the $A_1$ is up to the mark. Performance of $A_2$ is superior to $A_3$. The study proposes that PPI on Covid-19 candidate gene is extremely crucial for human disease.

%\begin{figure}%[!htb]
%	\minipage{0.50\textwidth}
%	\includegraphics[width=\linewidth]{204_PPI_Network_G1.png}
%	\caption{PPI Network $G_1$ }\label{PPI_Network_G1}
%	\endminipage\hfill
%	\minipage{0.50\textwidth}
%	\includegraphics[width=\linewidth]{254_PPI_Network_G2.png}
%	\caption{PPI Network $G_2$ }\label{PPI_Network_G2}
%	\endminipage
	
	%
%\end{figure}
\vspace{-9mm}
\vspace{-14mm}
A higher $DI$ indicates a better cluster. Dunn index is defined in terms of inter-cluster distance and intra-cluster distance. Minimum intra-cluster distance and maximum inter-cluster distance are the criteria for DI.

\clearpage

\vspace{-14mm}
\vspace{-14mm}
\vspace{-14mm}
The analysis shows that the performance of $A_1$ is good enough and $A_2$ performs better than $A_3$. In a randomly generated geometric network, random points are generated and placed with 2 dimension unit cube. We have taken distance threshold as the radius of 0.3 and distance metric as $2$. 

\clearpage
\begin{figure}[H]{}
	\begin{minipage}{0.3\textwidth}
		\begin{center}
			\begin{tikzpicture}[scale=0.5]{} 
				\begin{axis}[
					xlabel=$\overset{\rightarrow}{i}$,
					ylabel=$\overset{\rightarrow}{t}$,]
					\addplot[smooth,mark=*,blue] plot coordinates {
					(0, 0.050086)
					(1, 0.033876)
					(2, 0.039767)
					(3, 0.035909)
					(4, 0.029606)
					(5, 0.031948)
					(6, 0.03681)
					};
					%\addlegendentry{Algorithm 1}
					
					\addplot[smooth,color=red,mark=x]
					plot coordinates {
						(0, 0.037403)
						(1, 0.048866)
						(2, 0.040342)
						(3, 0.028112)
						(4, 0.03885)
						(5, 0.025245)
						(6, 0.03672)
					};
					%\addlegendentry{Algorithm 2}
					
					\addplot[smooth,color=green,mark=x]
					plot coordinates {
						(0, 0.039541)
						(1, 0.037101)
						(2, 0.038138)
						(3, 0.025413)
						(4, 0.035588)
						(5, 0.035494)
						(6, 0.02891)
					};
					%\addlegendentry{Algorithm 3}
				\end{axis}
			\end{tikzpicture}\caption{E of $G_1$}\label{E_G1}
			
		\end{center}
	\end{minipage}\hfillx
	\begin{minipage}{0.3\textwidth}
		\begin{center}
			
			\begin{tikzpicture}[scale=0.5]{}
				\begin{axis}[
					xlabel=$\overset{\rightarrow}{i}$,
					ylabel=$\overset{\rightarrow}{t}$,
					legend style={at={(2.45,0.8)},
						anchor=north}]
					\addplot[smooth,mark=*,blue] plot coordinates {
						(0, 0.091521)
						(1, 0.051931)
						(2, 0.039953)
						(3, 0.039195)
						(4, 0.042476)
					};
					\addlegendentry{$A_1$}
					
					\addplot[smooth,color=red,mark=x]
					plot coordinates {
						(0, 0.057538)
						(1, 0.049052)
						(2, 0.038865)
						(3, 0.039662)
						(4, 0.027268)
						(5, 0.022966)
						(6, 0.026573)
						(7, 0.034874)
						
					};
					\addlegendentry{$A_2$}
					
					\addplot[smooth,color=green,mark=x]
					plot coordinates {
						(0, 0.043338)
						(1, 0.040875)
						(2, 0.037385)
						(3, 0.038825)
						(4, 0.036317)
						(5, 0.036827)
						(6, 0.036894)
					};
					\addlegendentry{$A_3$}
				\end{axis}
			\end{tikzpicture}
			\caption{E of $G_2$}\label{E_G2}
		\end{center}
	\end{minipage}\hfillx
	\begin{minipage}{0.3\textwidth}
		\begin{center}		
			\begin{tikzpicture}[scale=0.5]
				\begin{axis}[
					xlabel=$\overset{\rightarrow}{i}$,
					ylabel=$\overset{\rightarrow}{t}$,]
					\addplot[smooth,mark=*,blue] plot coordinates {
						(0, 0.132476)
						(1, 0.142994)
						(2, 0.120654)
						(3, 0.096694)
						(4, 0.118854)
						(5, 0.128957)
						(6, 0.13863)
						(7, 0.102777)
					};
					%\addlegendentry{Algorithm 1}
					
					\addplot[smooth,color=red,mark=x]
					plot coordinates {
						(0, 0.048619)
						(1, 0.031656)
						(2, 0.033559)
						(3, 0.027367)
						(4, 0.038709)
						(5, 0.017767)
						(6, 0.018253)
						(7, 0.011587)
						(8, 0.03622)
						(9, 0.01258)
					};
					%\addlegendentry{Algorithm 2}
					
					\addplot[smooth,color=green,mark=x]
					plot coordinates {
					(0, 0.046492)
					(1, 0.031935)
					(2, 0.030568)
					(3, 0.01424)
					(4, 0.017639)
					(5, 0.009668)
					(6, 0.024644)
					(7, 0.028035)
					};
					%\addlegendentry{Algorithm 3}
				\end{axis}
			\end{tikzpicture}\caption{E of $G_3$ }\label{E_G3}
		\end{center}
	\end{minipage}
\end{figure}\vspace{-15mm}
\begin{figure}[H]\hypertarget{eg1}{}
	\begin{minipage}{0.3\textwidth}
		\begin{center}
			\begin{tikzpicture}[scale=0.5] \label{runTG1}
				\begin{axis}[
					xlabel=$\overset{\rightarrow}{i}$,
					ylabel=$\overset{\rightarrow}{t}$,]
					\addplot[smooth,mark=*,blue] plot coordinates {
						(0, 0.083066)
						(1, 0.189443)
						(2, 0.065712)
						(3, 0.049373)
						(4, 0.035142)
						(5, 0.025696)
						(6, 0.024826)
						(7, 0.039916)
					};
					%\addlegendentry{Algorithm 1}
					
					\addplot[smooth,color=red,mark=x]
					plot coordinates {
					(0, 0.044147)
					(1, 0.056912)
					(2, 0.036313)
					(3, 0.041602)
					(4, 0.038629)
					(5, 0.037917)
					(6, 0.040117)
					(7, 0.03396)
					(8, 0.024347)
					(9, 0.02849)
					};
					%\addlegendentry{Algorithm 2}
					
					\addplot[smooth,color=green,mark=x]
					plot coordinates {
						(0, 0.051619)
						(1, 0.04519)
						(2, 0.035982)
						(3, 0.039129)
						(4, 0.03885)
						(5, 0.029184)
						(6, 0.036769)
						(7, 0.03862)
					};
					%\addlegendentry{Algorithm 3}
				\end{axis}
			\end{tikzpicture}\caption{E of $G_4$}\label{E_G4}
			
		\end{center}
	\end{minipage}\hfillx
	\begin{minipage}{0.3\textwidth}
		\begin{center}
			
			\begin{tikzpicture}[scale=0.5]
				\begin{axis}[
					xlabel=$\overset{\rightarrow}{i}$,
					ylabel=$\overset{\rightarrow}{S_p\%}$,]
					legend style={at={(2.45,0.8)},
						anchor=north}]
					\addplot[smooth,mark=*,blue] plot coordinates {
						(0, 5.886214994570844)
						(1, 11.39870205297846)
						(2, 3.4771849195727382)
						(3, 1.911567889699755)
						(4, 1.4014797606121057)
						(5, 1.2954218327820004)
						(6, 1.2954218327820004)
						(7, 0.5025125628140703)
					};
					\addlegendentry{$A_1$}
					
					\addplot[smooth,color=red,mark=x]
					plot coordinates {
						(0, 5.886214994570844)
						(1, 21.125729148253832)
						(2, 3.7322289841165626)
						(3, 1.608545238756597)
						(4, 0.6464483220120704)
						(5, 0.5227140728769476)
						(6, 0.50503775157193)
						(7, 0.5025125628140703)
						
					};
					\addlegendentry{$A_2$}
					
					\addplot[smooth,color=green,mark=x]
					plot coordinates {
					(0, 5.886214994570844)
					(1, 5.795308199287897)
					(2, 1.8231862831746672)
					(3, 0.6464483220120704)
					(4, 0.5151385066033686)
					(5, 0.50503775157193)
					(6, 0.5025125628140703)
					(7, 0.5025125628140703)
					};
					\addlegendentry{$A_3$}
				\end{axis}
			\end{tikzpicture}
			\caption{S of $G_1$}\label{S_G1}
		\end{center}
	\end{minipage}\hfillx
	\begin{minipage}{0.3\textwidth}
		\begin{center}		
			\begin{tikzpicture}[scale=0.5]
				\begin{axis}[
				xlabel=$\overset{\rightarrow}{i}$,
				ylabel=$\overset{\rightarrow}{S_p\%}$,]
					\addplot[smooth,mark=*,blue] plot coordinates {
						(0, 4.5623048907388135)
						(1, 7.310093652445369)
						(2, 2.778681061394381)
						(3, 1.5283558792924037)
						(4, 0.4032258064516129)
						(5, 0.4032258064516129)
					};
					%\addlegendentry{Algorithm 1}
					
					\addplot[smooth,color=red,mark=x]
					plot coordinates {
						(0, 4.5623048907388135)
						(1, 16.43632934443288)
						(2, 2.7770551508844954)
						(3, 1.1820369406867846)
						(4, 0.4893990634755463)
						(5, 0.4292403746097815)
						(6, 0.406477627471384)
						(7, 0.4032258064516129)
						(8, 0.4032258064516129)
					};
					%\addlegendentry{Algorithm 2}
					
					\addplot[smooth,color=green,mark=x]
					plot coordinates {
						(0, 4.5623048907388135)
						(1, 4.188345473465141)
						(2, 1.4373048907388137)
						(3, 0.518665452653486)
						(4, 0.41948491155046824)
						(5, 0.406477627471384)
						(6, 0.4032258064516129)
						(7, 0.4032258064516129)
					};
					%\addlegendentry{Algorithm 3}
				\end{axis}
			\end{tikzpicture}\caption{S of $G_2$ } \label{S_G2}
		\end{center}
	\end{minipage}
\end{figure}\vspace{-5mm}
\begin{figure}[H]\hypertarget{eg1}{}
	\begin{minipage}{0.3\textwidth}
		\begin{center}
			\begin{tikzpicture}[scale=0.5] 
				\begin{axis}[
				xlabel=$\overset{\rightarrow}{i}$,
				ylabel=$\overset{\rightarrow}{S_p\%}$,]
					\addplot[smooth,mark=*,blue] plot coordinates {
						(0, 6.935)
						(1, 20.6875)
						(2, 17.045)
						(3, 6.6175)
						(4, 1.3425)
						(5, 0.765)
						(6, 0.72)
						(7, 0.61)
						(8, 0.5)
					};
					%\addlegendentry{Algorithm 1}
					
					\addplot[smooth,color=red,mark=x]
					plot coordinates {
						(0, 6.935)
						(1, 12.185)
						(2, 8.045)
						(3, 4.455)
						(4, 1.7375)
						(5, 0.9525)
						(6, 0.655)
						(7, 0.585)
						(8, 0.5275)
						(9, 0.5275)
						(10, 0.5225)
					};
					%\addlegendentry{Algorithm 2}
					
					\addplot[smooth,color=green,mark=x]
					plot coordinates {
						(0, 6.935)
						(1, 7.63)
						(2, 2.5775)
						(3, 0.9575)
						(4, 0.6025)
						(5, 0.5425)
						(6, 0.54)
						(7, 0.5375)
						(8, 0.5325)
					};
					%\addlegendentry{Algorithm 3}
				\end{axis}
			\end{tikzpicture}\caption{S of $G_3$}\label{S_G3}
			
		\end{center}
	\end{minipage}\hfillx
	\begin{minipage}{0.3\textwidth}
		\begin{center}
			
			\begin{tikzpicture}[scale=0.5]{}
				\begin{axis}[
					xlabel=$\overset{\rightarrow}{i}$,
					ylabel=$\overset{\rightarrow}{S_p\%}$,]
					legend style={at={(2.45,0.8)},
						anchor=north}]
					\addplot[smooth,mark=*,blue] plot coordinates {
						(0, 6.448)
						(1, 18.6448)
						(2, 13.1104)
						(3, 4.9808)
						(4, 1.296)
						(5, 0.712)
						(6, 0.4672)
						(7, 0.4)
						(8, 0.4)
					};
					\addlegendentry{$A_1$}
					
					\addplot[smooth,color=red,mark=x]
					plot coordinates {
						(0, 6.448)
						(1, 11.2416)
						(2, 7.4736)
						(3, 4.4528)
						(4, 1.8448)
						(5, 0.8208)
						(6, 0.4864)
						(7, 0.4224)
						(8, 0.408)
						(9, 0.4064)
						(10, 0.4048)
						
					};
					\addlegendentry{$A_2$}
					
					\addplot[smooth,color=green,mark=x]
					plot coordinates {
						(0, 6.448)
						(1, 7.0144)
						(2, 2.3456)
						(3, 0.7824)
						(4, 0.464)
						(5, 0.4256)
						(6, 0.4224)
						(7, 0.4208)
						(8, 0.4192)
					};
					\addlegendentry{$A_3$}
				\end{axis}
			\end{tikzpicture}
			\caption{S of $G_4$}\label{S_G4}
		\end{center}
	\end{minipage}\hfillx
	\begin{minipage}{0.3\textwidth}
		\begin{center}		
				\begin{tikzpicture}[scale=0.50]
				\begin{axis}[xlabel=$\overset{\rightarrow}{f}$,
					ylabel=$\overset{\rightarrow}{n_c}$,
					ymin=0, ymax=55,
					minor y tick num = 3,
					area style,
					]
					\addplot+[ybar interval,mark=no] plot coordinates {(1, 36)
						(2, 23)
						(3, 11)
						(4, 4)
						(5, 6)
						(6, 5)
						(7, 2)
						(8, 1)
						(9, 3)
						(10, 1)};
				\end{axis}
			\end{tikzpicture}\caption{H of $G_4$ on $A_3$ } \label{Histo_G3A3}
		\end{center}
	\end{minipage}
\end{figure}\vspace{-5mm}
\begin{figure}[H]
	\begin{minipage}{0.3\textwidth}
		\begin{center}
		\begin{tikzpicture}[scale=0.50]
			\begin{axis}[xlabel=$\overset{\rightarrow}{f}$,
				ylabel=$\overset{\rightarrow}{n_c}$,
				ymin=0, ymax=55,
				minor y tick num = 3,
				area style,
				]
				\addplot+[ybar interval,mark=no] plot coordinates {(1, 28)
					(2, 14)
					(3, 10)
					(4, 13)
					(5, 1)
					(6, 1)
					(7, 2)
					(8, 1)
					(9, 1)
					(10, 2)};
			\end{axis}
		\end{tikzpicture}\caption{H of $G_4$ on $A_3$ } \label{Histo_G4A3}

		\end{center}
	\end{minipage}\hfillx
	\begin{minipage}{0.3\textwidth}
		\begin{center}
			\begin{tikzpicture}[scale=0.5]
				\begin{axis}[
					xlabel=$\overset{\rightarrow}{i}$,
					ylabel=$\overset{\rightarrow}{r_i}$,legend style={at={(2.45,0.8)},
						anchor=north}]
					\addplot[smooth,mark=*,blue] plot coordinates {
						(150, 3.556560893574651)
						(175, 2.6791072696499225)
						(200, 3.4416224766316)
						(225, 1.8943230389454193)
						(250, 4.060939764827224)
						
					};
					\addlegendentry{$A_1$}
					
					\addplot[smooth,color=red,mark=x]
					plot coordinates {
						(150, 0.13841438271754963)
						(175, 0.19584902242558452)
						(200, 0.25584003927183835)
						(225, 0.25327114688353936)
						(250, 0.1505808729995455)
					};
					\addlegendentry{$A_2$}
					
					\addplot[smooth,color=green,mark=x]
					plot coordinates {
						(150,0.10941170885741706)
						(175, 0.05221765918579668)
						(200, 0.1695257406938393)
						(225, 0.07396585876295046)
						(250, 0.05126178914266123)
					};
					\addlegendentry{$A_3$}
				\end{axis}
			\end{tikzpicture}\caption{DI on random} \label{DI_random}
			
		\end{center}
	\end{minipage}\hfillx
	\begin{minipage}{0.3\textwidth}
		\begin{center}
			\begin{tikzpicture}[scale=0.5]
				\begin{axis}[
					xlabel=$\overset{\rightarrow}{i}$,
					ylabel=$\overset{\rightarrow}{r_i}$,]
					\addplot[smooth,mark=*,blue] plot coordinates {
						(152, 41.924055102355624)
						(162, 56.8067870054531)
						(171, 66.30357147245364)
						(190, 163.27189333196287)
						(209, 59.377919758942916)
						(229, 61.48382354868521)
						(248, 130.89526948989098)
					};
					%\addlegendentry{$A_1$}
					
					\addplot[smooth,color=red,mark=x]
					plot coordinates {
						(151, 32.62588036434948)
						(162, 14.000695011492194)
						(171, 28.03582592896932)
						(190, 39.721951765230045)
						(209, 24.428972967347455)
						(229, 38.187405572563215)
						(248, 41.698478349233554)

					};
					%\addlegendentry{$A_2$}
					
					\addplot[smooth,color=green,mark=x]
					plot coordinates {
						(152, 11.190581482915826)
						(162, 10.734333155180895)
						(171, 24.12781725539563)
						(190, 13.78559966038304)
						(209, 23.489276266255345)
						(229, 11.035465254326121)
						(248, 17.037557456396623)
					};
					%\addlegendentry{$A_3$}
				\end{axis}
			\end{tikzpicture}\caption{DI on PPI} \label{DI_PPI}
		\end{center}
	\end{minipage}
\end{figure}\vspace{-15mm}
\clearpage

\section{Discussion and Conclusions}\vspace{-4mm}
In this paper, we propose graph-based clustering by the help of MCL, RMCL, and MCL with a variable inflation rate algorithm. We have validated our clustering using DI and the quality of the clustering is very magnificent. To evaluate our clustering algorithm, the DI metric is interpreted as an intra-cluster and inter-cluster distance. DI in the results section shows that the performance of the $A_1$ is up to the mark. Performance of $A_2$ is superior to $A_3$. The study proposes that PPI on the Covid-19 candidate gene is extremely crucial for human disease.\vspace{-5mm}

%Table~\ref{tab1} gives a summary of all heading levels.

%\noindent Displayed equations are centered and set on a separate line.
%\begin{equation}
%x + y = z
%\end{equation}
%Please try to avoid rasterized images for line-art diagrams and schemas. Whenever possible, use vector graphics instead (see Fig.~\ref{fig1}).

%
% ---- Bibliography ----
%
% BibTeX users should specify bibliography style 'splncs04'.
% References will then be sorted and formatted in the correct style.
%
\bibliographystyle{splncs04}
\bibliography{mybibliography}
%
%\begin{thebibliography}{8}
%\bibitem{ref_article1}

%\end{thebibliography}
\end{document}